%% file: FicNet1026.tex
\documentclass[final, twoside]{IEEEtran} 
\usepackage{float}
\usepackage{tikz}
\usepackage{amssymb, amsmath, amsthm, bm}
\usepackage{xcolor}
\usepackage{pgfplots}

\usepackage{subfigure}
\pgfplotsset{compat=newest}
\usepackage[normalem]{ulem}
\usepackage{ulem}

\makeatletter
\newcommand{\tpmod}[1]{{\@displayfalse\pmod{#1}}}
\makeatother

\usepackage{graphicx, color}
\graphicspath{{figures-pdf/}}

\usepackage{cite}

\usepackage{url}
\usepackage{diagbox}

\usepackage{threeparttable}

\usepackage[mathlines]{lineno}

\usepackage[linesnumbered, ruled]{algorithm2e}

\usepackage{booktabs}

\usepackage{algpseudocode}
\usepackage{multirow, bigstrut}
\usepackage{tabularx}
\usepackage{arydshln}
\usepackage{empheq}
\usepackage{datetime}

\newlength\OneImW
\newlength\BigOneImW
\newlength\twofigwidth
\newlength\sfigwidth
\newlength\vfigskip
\newlength\figsep
\usepackage[bookmarks=false]{hyperref}
\hypersetup{linktocpage=true, pdfborderstyle={/S/S/W 1}}

\newcommand{\newItemizeWidth}{%
	\setlength{\labelwidth}{\widthof{\textbullet}}%
	\setlength{\labelsep}{1pt}%
	\setlength{\IEEElabelindent}{0pt}%
	\IEEEiedlabeljustifyl
}

\usepackage{textcomp}
\definecolor{mycolor1}{rgb}{1.00000,0.00000,1.00000}
\begin{document}
	
\title{
Few-shot Fine-grained Image Classification via Multi-Frequency Neighborhood and Double-cross Modulation}

\author{
Hegui Zhu,
Zhan Gao,
Jiayi Wang,
Yange Zhou,
Chengqing Li
\thanks{This work was funded by the Natural Science Foundation of Liaoning Province (No.~2020-MS-080), the National Key Research and Development Program of China (No.~2017YFF0108800). (Corresponding author: Chengqing Li)}

\thanks{H. Zhu, Z. Gao, J. Wang and Y. Zhou are with the Department of Mathematics, College of Sciences, Northeastern University, Shenyang 110819, Liaoning, China (e-mail: zhuhegui@mail.neu.edu.cn).}

\thanks{C. Li is with MOE (Ministry of Education) Key Laboratory of Intelligent Computing and Information Processing, Xiangtan University, Xiangtan 411105, Hunan, China.}
}

\markboth{IEEE Transactions}{Zhu \MakeLowercase{et al.}}
\IEEEpubid{\begin{minipage}{\textwidth}\ \\[12pt] \centering
			1520-9210 \copyright 2020 IEEE. Personal use is permitted, but republication/redistribution requires IEEE permission.\\
			See https://www.ieee.org/publications/rights/index.html for more information. \\
			\today
	\end{minipage}
}

\maketitle

\begin{abstract}
Traditional fine-grained image classification typically relies on large-scale training samples with annotated ground-truth. However, some sub-categories have few available samples in real-world applications, and current few-shot models still have difficulty in distinguishing subtle differences among fine-grained categories. To solve this challenge,
we propose a novel few-shot fine-grained image classification network (FicNet) using multi-frequency neighborhood (MFN) and double-cross modulation (DCM). MFN focuses on both spatial domain and frequency domain to capture multi-frequency structural representations, which reduces the influence of appearance and background changes to the intra-class distance. DCM consists of bi-crisscross component and double 3D cross-attention component. It modulates the representations by considering global context information and inter-class relationship respectively, which enables the support and query samples respond to the same parts and accurately identify the subtle inter-class differences. The comprehensive experiments on three fine-grained benchmark datasets for two few-shot tasks verify that FicNet has excellent performance compared to the state-of-the-art methods. Especially, the experiments on two datasets, ``Caltech-UCSD Birds" and ``Stanford Cars", can obtain classification accuracy 93.17\% and 95.36\%, respectively. They are even higher than that the general fine-grained image classification methods can achieve.
\end{abstract}

\section{Introduction}

\begin{figure}[!htb]
\centering
\includegraphics[width=\BigOneImW]{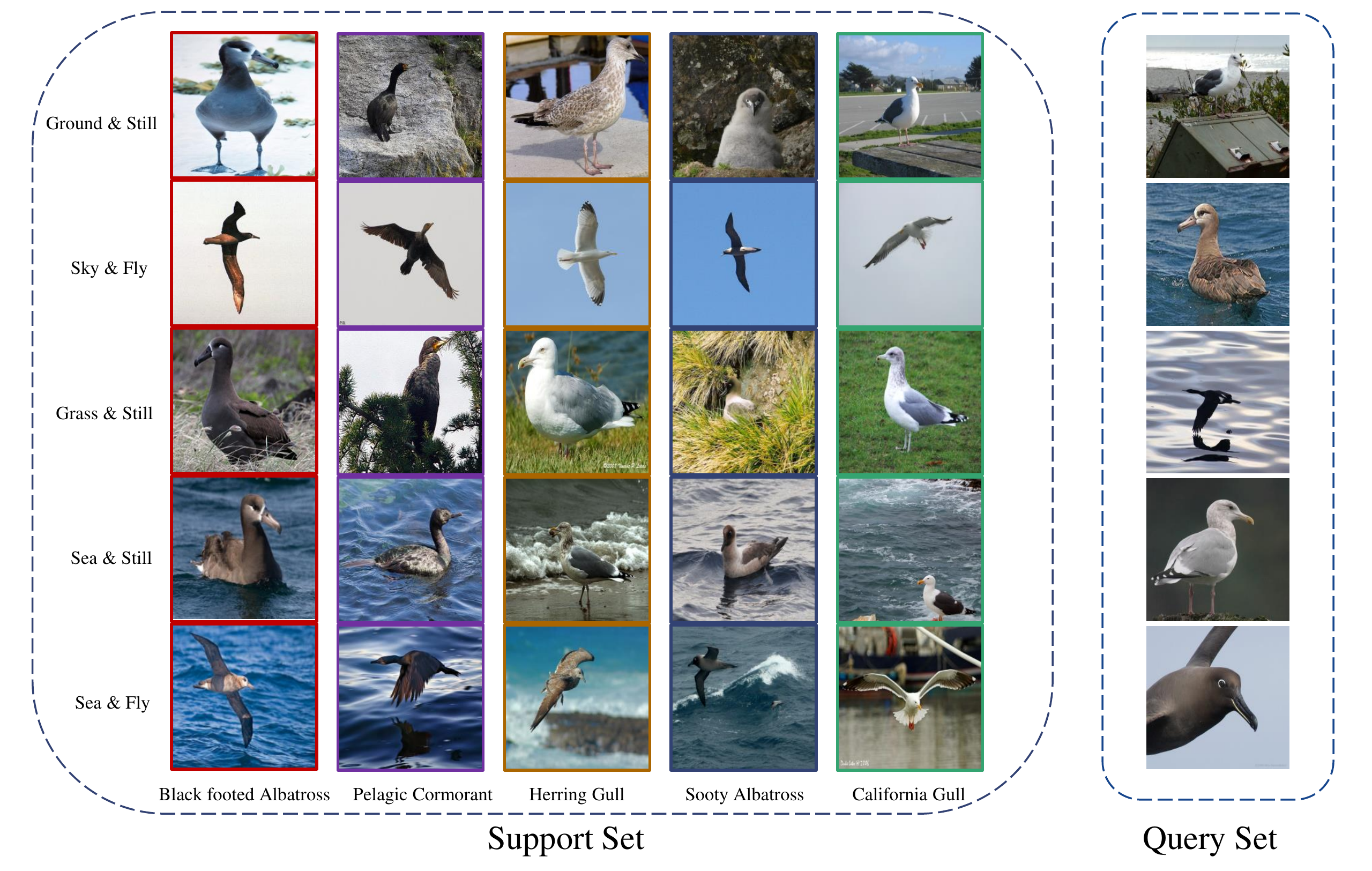}
\caption{Illustration of a ``5-way 5-shot" FSFG task.
Different columns represent different bird species: ``Black footed Albatross", ``Pelagic Cormorant", ``Herring Gull", ``Sooty Albatross", ``California Gull".
Different rows represent different poses, ``still" and ``fly", and backgrounds, ``ground", ``sky", ``grass", and ``sea".}
\label{fig.FSFG}
\end{figure}

\IEEEPARstart{F}ine-grained image classification aims to accurately distinguish different sub-categories belonging to the same entry-level category, such as different kinds of birds, flowers, cars. Particularly, with the development of deep learning \cite{zhou:CS:TM21, cqli:CS:TMM22}, some fine-grained classification methods based on neural network have attracted much attention in various application scenarios, such as automatic biodiversity monitoring \cite{van2021benchmarking} and intelligent detection \cite{guan2022lightweight}. Generally, fine-grained classification relies on a large number of labeled samples. However, there are only few training samples for some sub-categories in many fields, e.g. rare cases in medical research \cite{feng2021interactive},
unbalanced time series classification \cite{zhu:driver:TIM22}, new species in biological research \cite{sun2020few}. Because of the obvious technical merits, many few-shot learning (FSL) methods have been proposed to categorize the object images with few training samples in recent years \cite{vinyals2016matching, chen2020multi,Lee_2022_CVPR}. However, they cannot capture the subtle features to deal with fine-grained classification task. Therefore, solving the few-shot fine-grained (FSFG) image classification task attracted intensive attentions from research community.
There are still some improvement space on FSFG learning because of the following two problems:
1) the intra-class distances of sub-categories may be very large. The appearance and background is often a key factor to increase the inter-class distance, which plays a negative role in the FSFG task. As shown in Fig.~\ref{fig.FSFG}, even in the same row, the posture and background are very different;
2) the inter-class distances among different sub-categories may be small.
Note that only local subtle features can be used to distinguish different subcategories.
For different tasks, the features involved in the classification may be very different. In each column of Fig.~\ref{fig.FSFG}, only the beak or wings of birds are different, so it's best to pay attention to the beak or wings. Moreover, in the last query image of query set, there is only the head of the bird because of the angle of view and distance from the camera, so it's best to focus on the bird's head for both query and support images. All these increase the classification difficulty because of the subtle difference among different sub-categories.

\IEEEpubidadjcol 

For the first problem, it depends on the complex network structure and training strategy to obtain a robust feature of each class. Some methods based on feature enhancement\cite{wei2019piecewise,li2019distribution,huang2019compare} have been proposed to alleviate it. For example, Wei et al. \cite{wei2019piecewise} augmented single coding feature with a self-bilinear encoding network; Huang et al. \cite{huang2019compare} proposed pairwise bilinear pooling network to extract second-order descriptors for obtaining more precise feature representations. However, they regard the image as a whole space in terms of
pixel or feature to enhance features, which ignores the structure of the target object and cannot avoid the negative impact of the background.
For the second problem, Zhu et al. \cite{zhu2020multi} proposed a multi-attention method to get the discriminative parts of images; Huang et al. \cite{huang2020low} employed a new low-rank bilinear pool operation to capture subtle differences between support and query set. However, they do not consider that the features used to distinguish categories may vary significantly for different tasks, and the samples in the support set and query set may also respond to different parts of the target object. Therefore, it not only needs to obtain more portable and complete features of the target object, but also align the features in support set and query set.

In this paper, we propose a novel few-shot fine-grained image classification network (FicNet) using multi-frequency neighborhood (MFN) and double-cross modulation (DCM). MFN captures important relationships among the neighborhoods and uses self-similarity to extract its own smooth spatial structure, which can reduce the change sensitivity of intra-class appearance. Then it decomposes the structural features into multi-frequency components, which captures compact structural patterns to remove irrelevant features. DCM module consists of bi-crisscross (BCC) and double 3d cross-attention (DCA). BCC is used to extract more context information from different regions, which modulates the embedding to include more complete target information. DCA refines the correlation between the support and query samples to generate reliable co-attention map, which modulates the embedding to contain task-level discriminative features. Then the samples can respond to the same parts of the target object. The extensive experiments show that FicNet significantly outperforms existing methods on fine-grained benchmark datasets. Our technical contributions are summarized as follows:
\begin{itemize}
\item Proposing a novel and compact end-to-end network, which uses two branches of capture and modulation to solve FSFG classification pertinently.

\item Designing module MFN to fuse spatial and frequency information, which can capture multi-frequency structural representation to reduce the change influence of appearance and background to intra-class distance.

\item Constructing module DCM to modulate representation from both context and task-specific information, which makes the involved samples respond to the same parts and identifies subtle inter-class differences accurately.
\end{itemize}

The rest of the paper is organized as follows.
First, the related work on fine-grained image classification and associate learning methods is briefly reviewed in Sec.~\ref{sec.work}.
Then, the adopted methodology to design FicNet is presented in Sec.~\ref{sec:methodology}.
Section~\ref{sec:Experimental} presents detailed experimental settings and results to verify the effectiveness of FicNet. Furthermore, Sec.~\ref{sec:Ablation} provides extensive ablation studies on every basic part of FicNet. The last section concludes the paper.

\section{The related work}
\label{sec.work}

To demonstrate the development background of FicNet, we briefly review the related work on few-shot learning, fine-grained image classification, few-shot fine-grained learning and frequency domain learning.

\begin{itemize}[\newItemizeWidth]

\item \textbf{Few-shot learning}

FSL simulates human ability to acquire knowledge from a few samples through generalization and analogy. In recent years, many methods based on metric learning emerged. For example, Snell et al. \cite{snell2017prototypical} proposed a simple and efficient network ProtoNet for few-shot learning, which performed classification by calculating Euclidean distance from the prototype expression of each category. In \cite{zhang2022mlso}, Zhang et al. employed second-order relational descriptors from feature maps to capture the similarity between the support-query pairs. 
Finn et al. \cite{finn2017model} provided the most typical meta-learning method model-agnostic meta-learning algorithm (MAML), which can learn a set of good initialization parameters. Furthermore, Wang et al. \cite{Zhong2022graph} devised a graph complemented latent representation (GCLR) network to learn a better representation. To alleviate over-fitting phenomenon in FSL from estimation of data distribution, Yang et al. \cite{yang2021free} used calibrated distribution to provide more diverse generations
using statistical information from classes with abundant samples.

\item \textbf{Fine-grained image classification}

For many years, fine-grained image classification has been a hot topic in the field of computer vision. Deep fine-grained classification methods can be broadly divided into two categories: regional features-based methods and global features-based methods. Zhang et al. first combined R-CNN into a fine-grained classifier with geometric prior, and used object detection to locate local regions \cite{zhang2014part}. For easy deployment, Zhang et al. \cite{zhang2016weakly} provided a fine-grained image categorization system, which only classified labels for training images. Different from simply locating different parts, the approach given in \cite{zhang2019learning} attempted to mine complementary information from multiple granularity parts. The method designed in \cite{song2020bi} can not only capture distinguishing parts, but also use knowledge in text description for interactive alignment. In contrast, the global features-based methods extract features from the whole image without explicit localization of the target. As a typical work, Feng \cite{feng2022kernel} used kernel approximation to introduce high-order information, and then adopted statistical information to enhance modeling ability.

\item \textbf{Few-shot fine-grained learning}

Wei et al. \cite{wei2019piecewise} proposed the first FSFG model, which used self-bilinear encoder to capture subtle image features. Li et al. \cite{li2019distribution} further replaced the naive self-bilinear encoder with covariance pooling and designed covariance metric network, which measured the relationship between a query sample and each category by the distribution consistency. Different from these self-linear methods, Huang et al. \cite{huang2019compare} provided a novel pairwise-bilinear pooling method to compare subtle differences between fine-grained categories. Then they further employed a new low-rank bilinear pool operation \cite{huang2020low}, and designed a feature alignment layer to match the support features with the query features. Meanwhile, Wang et al. \cite{wang2021fine} proposed a data enhancement method FOT (Foreground Object Transformation), which used the posture transformation generator to generate additional samples of new sub-categories. Zhu et al. \cite{zhu2020multi} provided a multi-attention meta learning method that exploited the attention mechanism of base learner and task learner to capture the different parts of an image. Moreover, Ruan et al. \cite{ruan2021few} explored a spatial attentive comparison network (SACN), which consists of three separate modules for feature extraction, selective-comparison of similarity and classification.

\begin{figure*}[!htb]
\centering
\includegraphics[width=2\BigOneImW]{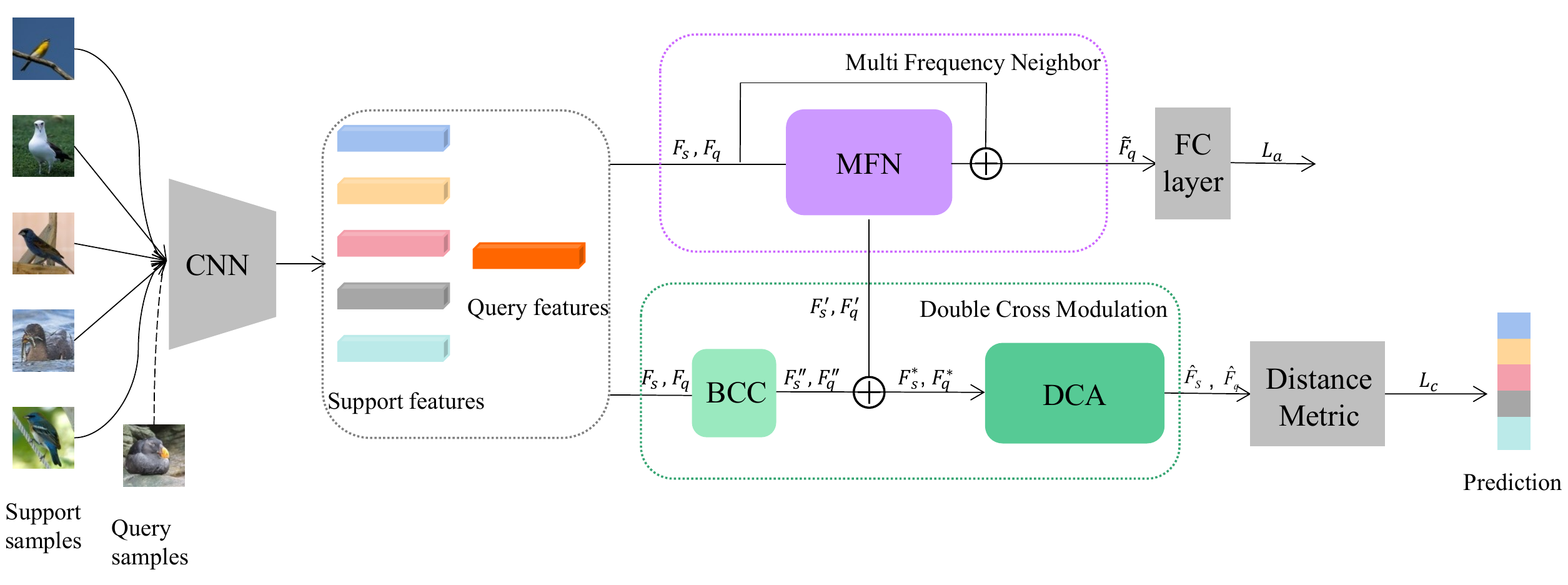}
\caption{The architecture of FicNet.}
\label{fig.2}
\end{figure*}

\begin{figure*}[!htb]
\centering
\includegraphics[width=2\BigOneImW]{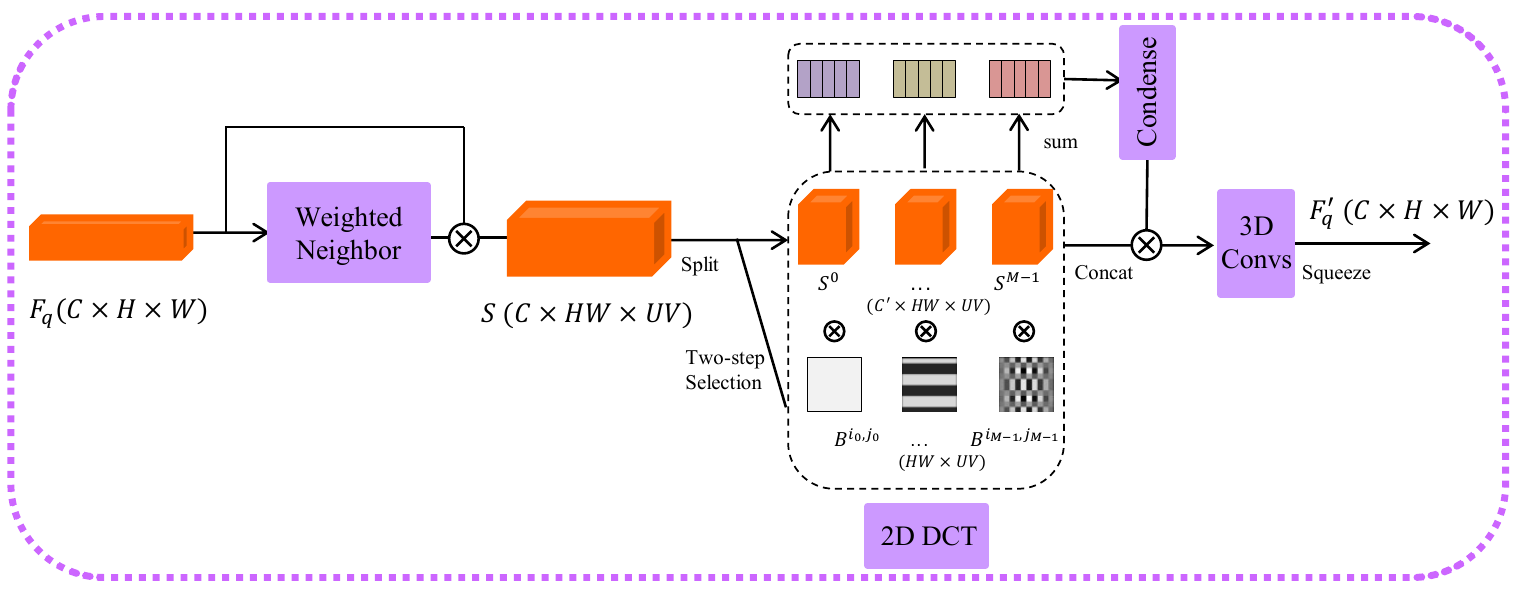}
\caption{The snapshot of module multi-frequency neighborhood.}
\label{fig.3}
\end{figure*}

\item \textbf{Frequency domain learning}

Ehrlich and Davis \cite{ehrlich2019deep} creatively proposed a model-conversion algorithm to convert CNN model from spatial domain to frequency domain. Furthermore, Wang et al. \cite{wang2022ft} directly exploited the frequency modality as complementary information and capture the local contextual incoherence to enhance boundary consistency.
To avoid such complex model transition procedure, Xu et al. \cite{xu2020learning} analyzed the spectral bias and provided a frequency selection method to identify the trivial frequency components. Meanwhile, Qin et al.\cite{qin2021fcanet} considered the channel representation problem as a compression process using frequency analysis. They proved that the conventional global average pooling is a special case of feature decomposition in the frequency domain.
\end{itemize}

Different from the previous image classification methods, FicNet tries to solve FSFG problem with capture branch and modulation branch. First, it captures rich and compact structural patterns to reduce the impact of appearance variation to intra-class distance by fusing spacial and frequency information. Then it modulates the embedding process to contain more complete class-related details based on global context and task-specific information. It can accurately identify the subtle inter-class differences.

\section{The adopted methodology}
\label{sec:methodology}

\subsection{Problem definition}
\label{sec:definition}

A FSFG problem is usually formalized as a ``$N$-way $K$-shot" task, and the dataset is divided into two parts:
training set $D_{\rm{tr}}=\{(I_i, y_i), y_i \in C^s\}_{i=1}^{N_{\rm{tr}}}$ and testing set $D_{\rm test}= \{(I_j, y_j), y_j\in C^t\}_{j=1}^{N_{\rm test}}$, where $I_i$ is the $i$-th fine-grained sample in $D_{\rm{tr}}$ and $y_i$ is its class label, $I_j$ is the $j$-th fine-grained sample in $D_{\rm test}$ and $y_j$ is its class label; the label sets satisfy $C^{s} \cap C^{t}=\varnothing$, $N_{\rm test}$ is the number of categories in test set.
We organize experiments in an episodic meta-training setting. Considering that both $D_{\rm tr}$ and $D_{\rm test}$ consist of many episodes, we randomly select $N$ different categories in each episode, and randomly choose $K$ and $P$ images in each category. They constitute support set $S=\{(I_s, y_s)\}_{s=1}^{\mathit{NK}}$ and query set $Q=\{(I_q, y_q)\}_{q=1}^{\mathit{NP}}$, respectively.
Every episode is iteratively sampled from $D_{\rm{tr}}$ to train FicNet.

\subsection{Architecture overview}
\label{ssec:Architecture}

FicNet belongs to the metric-based methods, and the specific structure is shown in Fig.~\ref{fig.2}, which is composed of two main modules, multi-frequency neighborhood (MFN) and double-cross modulation (DCM).
Note that marker \textcircled{+} means element-wise sum, $L_a$ denotes the auxiliary loss function, and $L_c$ represents the classification loss. As for the input samples, we employ ConvNet-4 as the feature extractor to provide basic feature maps, which are fed into MFN and converted into multi-frequency structural representations by analyzing the information in spatial and frequency domain. Then, BCC efficiently generates global context information to modulate multi-frequency structural representations, and DCA uses multiple 3D crossed convolution to perform modulation according to the current task.

\subsection{Multi-frequency neighborhood}
\label{sec3.2}

The module MFN is employed to capture compact multi-frequency structural representations by combining
the information in spatial and frequency domain, which can reduce the change sensitivity of intra-class representations. It includes weighted neighborhood and multi-frequency components. Figure~\ref{fig.3} illustrates the overall structure of MFN, where
$\otimes$ represents the Hadamard product. The details of MFN are introduced as follows.
\begin{itemize}[\newItemizeWidth]

\item \textbf{Weighted neighborhood}

Weighted neighborhood encodes local invariant structure by associating with neighborhoods of each feature in spatial domain. Because the operations on the support image and the query image are the same, so only the operations of query image are illustrated here.

Given a query image $I_q$, we employ ConvNet-4 to extract basic feature map $F_q \in R^{C \times H \times W}$. As for each feature $x$ in $F_q$, we extract its $UV-1$ neighbors through the sliding window, where $UV$ is the size of sliding block. Note that the nearer neighbor points obviously have a greater impact on $x$. So, one can
obtain the different weighted neighbor representation by
\begin{equation*}
F_{q*}(x+e)=\frac{1}{\Vert e \Vert+1} F_q(x+e),
\end{equation*}
where $x\in [0, H ] \times [0, W]$, $e \in [\frac{-U+1}{2}, \frac{U-1}{2}] \times [\frac{-V+1}{2}, \frac{V-1}{2}]$.
Then, one builds local spatial representation $s\in R^{C \times H \times W \times U \times V}$ by aggregating information of each $x$ in $F_q$ and its weighted neighbors $F_{q*}$:
\[
s(x, e) = \frac{F_q(x)}{ \|F_q(x)\|} \otimes \frac{F_{q*}(x+e)}{\|F_{q*}(x+e) \|}.
\]

\item \textbf{Multi-frequency components}

To further optimize the local spatial representation $s$, we convert it to frequency domain and represent it with frequency components. First, the local spatial representation $s$ is reshaped into $S \in R^{C \times HW \times UV}$, and equally divided into $M$ blocks along the channel dimension
\[
S=[S^0, S^1, \cdots, S^{M-1}],
\]
where channel dimension of each part $S^m$ is $\frac{C}{M}$.
Second, a two-dimensional discrete cosine transform (2D-DCT) is employed on $S^m$:
\begin{equation*}
\begin{aligned}
\mathit{Freq}^m & = 2 \mathrm{D}\text{-}\mathrm{DCT}^{i_m, j_m}(S^m) \\
	 & = \sum_{h=0}^{H-1} \sum_{w=0}^{W-1} S_{:, h, w}^m B_{h, w}^{i_m, j_m},
\end{aligned}
\end{equation*}
where $m\in\{0, 1, \cdots, M-1\}$, and $(i_m, j_m)$ is the index of the frequency component corresponding to $\mathit{Freq}^m$. In particular, the basis function $B_{h, w}^{i_m, j_m}$ is a pre-calculated constant, so this process does not introduce parameters. Moreover, the selection of index $(i_m, j_m)$ is also a key issue, and its detailed procedure is given as follows:
\begin{itemize}[\newItemizeWidth]
\item \textbf{Step~1} For each possible alternative frequency component corresponding to index $(m_i, m_j)$, evaluate the performance of the model using each component separately, where $m_i\in \{0, 1, \cdots, H-1\}$, $m_j\in\{0, 1, \cdots, W-1\}$. If a frequency component can bring higher classification accuracy, it is considered to be more important.

\item \textbf{Step~2} Produce the top-$M$ frequency components according to the importance demonstrated in \textbf{Step~1}: $\{(i_1, j_1), \cdots, (i_{M-1}, j_{M-1})\}$.
\end{itemize}

Then one integrates the scattered frequency information to get multi-frequency feature by
\begin{equation}
D=\operatorname{Sigmoid}(\operatorname{fc}([\mathit{Freq}^0, \mathit{Freq}^1, \cdots, \mathit{Freq}^{M-1}])),
\end{equation}
where $\operatorname{fc}$ is the fully connected layer, and
activation function $\operatorname{Sigmoid}(x)=\frac{e^x}{e^x+1}$.
Finally, we obtain the multi-frequency structural representation
\begin{equation*}
F'_q=\operatorname{Conv_{3d}}(D\otimes S),
\end{equation*}
where $\operatorname{Conv_{3d}}$ represents the 3D convolution for dimension reduction.
\end{itemize}

\begin{figure}[!htb]
\centering
	\includegraphics[width=\BigOneImW]{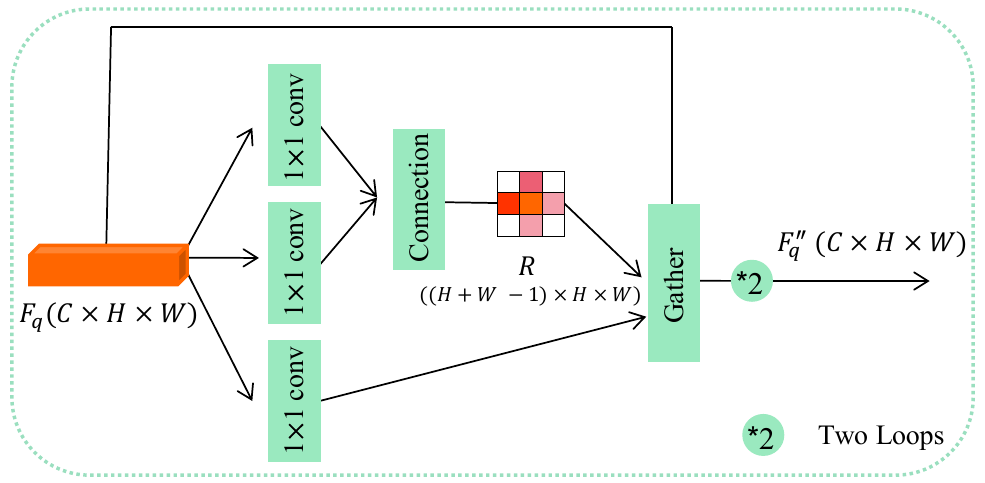}
	\caption{The structure of BCC component.}
	\label{fig.4}
\end{figure}

\begin{figure}[!htb]
\centering
 \subfigcapskip=-12pt 
 \subfigure[one]{
 \includegraphics[width=0.4\linewidth]{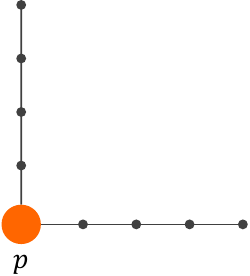}}
 \hspace{0cm}
 \subfigure[two]{
 \includegraphics[width=0.4\linewidth]{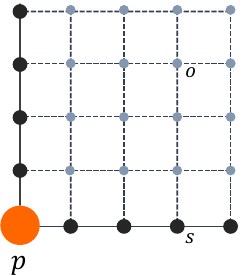}}
\vspace{-0.1cm}
\caption{Information flow of two loops in BCC.}
\label{infoflow}
\end{figure}

\begin{figure*}[!htb]
	\centering
	\includegraphics[width=2\BigOneImW]{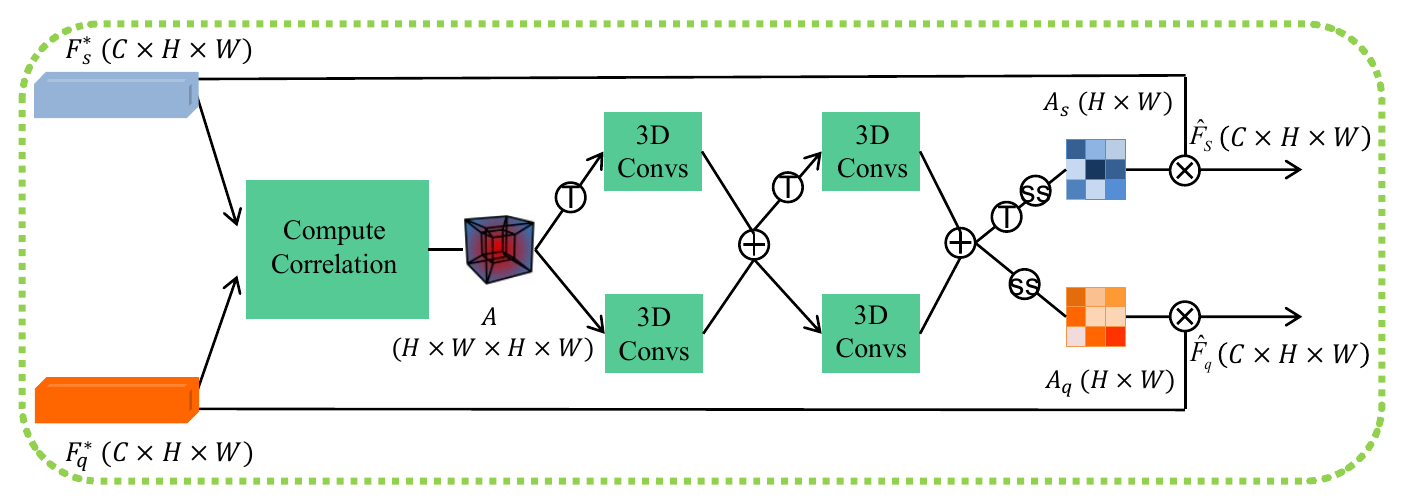}
	\caption{The structure of DCA, where \textcircled{ss} represents sum after Softmax, \textcircled{T} represents transpose.}
	\label{fig.5}
\end{figure*}

\subsection{Double-cross modulation}
\label{sec3.3}

To modulate multi-frequency structural representation $F'_q$ more complete and adapt to the current task, DCM is adopted to modulate feature from both context information and task-specific information. First, we use BCC to generate global context information, which is a supplement to $F'_q$. Then, DCA is used to adjust target region according to the current task.
\begin{itemize}[\newItemizeWidth]

\item \textbf{Bi-crisscross component}

Figure~\ref{fig.4} depicts the detailed contents of BCC. As for the given basic feature map $F_q \in R^{C \times H \times W}$ corresponding to query image $I_q$, we repeat the crisscross operation twice (represented by $*2$) on it. Different from self-attention \cite{Stand19}, there are three main differences. 1)Characteristics, BCC generates a sparse crisscross attention map for each position $p$ in $F_q$, which has only $H+W-1$ weights (orange grids), while self-attention \cite{Stand19} generates denser attention map and obtains local attention. 2) Information flow, the crisscross operation in BCC only gathers the context information of each position in the horizontal and vertical directions, and two same operations of sharing parameters can completely gather the information of all positions around it. While Self-attention\cite{Stand19}extracts a local region of pixels in each position with spatial extent $k$ centered around it, and gathers information of each position in a block. 3)Advantages, Our BCC can efficiently generate the global context information, which is adopted to modulate feature. While Self-attention\cite{Stand19} verifies that attention can be a stand-alone primitive to replace convolutions for vision model. The specific operation on $F_q$ is described as follows:
\begin{itemize}[\newItemizeWidth]
\item \textbf{Step~1} $F_q$ is fed into two $1 \times 1$ convolution layers to reduce channel dimension into $C''$ and a plain $1 \times 1$ convolution layer to generate three feature maps $Q$, $K$, and $V$, where $Q, K\in R^{C" \times H \times W}$ and $V \in R^{C \times H \times W}$. Denote feature vectors
of $K$ and $V$ in position $p$ as sets $\boldsymbol{\Omega}_p \in R^{(H+W-1) \times C''}$ and $\boldsymbol{\Phi}_p \in R^{(H+W-1) \times C}$, respectively.

\item \textbf{Step~2} Use connection operation to extract information in the horizontal and vertical directions by
\begin{equation}
R(i, p)=\operatorname{Softmax}(Q_p \boldsymbol{\Omega}_{i, p}^{\top}), \label{eq.9}
\end{equation}
where $\operatorname{Softmax}$ is the activation function $Q_p$ is a feature vector of $Q$ at position $p$, $\boldsymbol{\Omega}_{i, p} \in R^{C''}$ is the $i$-th element of $\boldsymbol{\Omega}_p$, $R(i, p)$ is a vector on $R \in R^{(H+W-1)\times H \times W}$ that reveals the degree of correlation between feature $Q_p$, $\boldsymbol{\Omega}_{i, p}$, $i\in\{1, 2, \cdots, H+W-1\}$.

\item \textbf{Step~3} Do the gather operation to aggregate the context information of each pixel in the cross path by
\begin{equation*}
\operatorname{gather}(R(i, p), \boldsymbol{\Phi}_p, F_q(p))=\sum_{i=1}^{H+W-1} R(i, p) \boldsymbol{\Phi}_{i, p}+F_q(p),
\end{equation*}
where $\boldsymbol{\Phi}_{i, p} \in R^{C''}$ is the $i$-th element of $\boldsymbol{\Phi}_p$, $F_q(p)$ is the feature vector of $F_q$ at position $p$.

\item \textbf{Step~4} Repeat the crisscross operation in Steps 1, 2, 3 twice and obtain $F''_q$. 
Figure~\ref{infoflow} presents intuitive explains for repeating and obtaining global context.
In loop one, position $p$ only gathers the information of horizontal and vertical features, and other positions do the same. In loop two, position $p$ can connect all other positions because the points which are vertical and horizontal to $p$ have completed the gather operation in loop one. For example, even if position $o$ is not in the same row or column with $p$, they can also be connected by other positions located in the same row or column with $p$,
such as position $s$.

\item \textbf{Step~5} Modulate $F'_q$ by employing $F''_q$ as the complement, and obtain the improved multi-frequency feature by
\begin{equation*}
F^*_q=F''_q+F'_q.
\end{equation*}
\end{itemize}
Note that the operations on the support samples are the same as that on the query samples.
The prototype $F^*_s$ of support samples $S$ can be obtained in a class-average manner.

\item \textbf{Double 3D cross-attention}

To further make modulation adapt to the current task, DCA takes a pair of support prototype $F^*_s$ and query $F^*_q$ to produce the corresponding cross-attention maps $A_s$ and $A_q$, respectively. Figure~\ref{fig.5} shows the pipeline of DCA.
First, calculate the correlation map $A\in R^{H \times W \times H \times W}$ with cosine distance, where the correlation degree of position $x_s$ in $F^*_s$ and $x_q$ in $F^*_q$ can be obtained by
\begin{equation*}
A(x_s, x_q)=\left(\frac{{F^*_s}(x_s)}{\|{F^*_s}(x_s)\|_2}\right)^\intercal
\left(\frac{{F^*_q}(x_q)}{\|{F^*_q}(x_q)\|_2}\right),
\end{equation*}
where $x_s, x_q\in [0, H]\times [0, W]$. Then use 3D double-cross convolutions to refine 4-D tensor $A$ with the following steps:
\begin{itemize}[\newItemizeWidth]
\item \textbf{Step~1} Reshape tensor $A$ from $R^{H\times W\times H\times W}$ into $R^{1\times H \times W\times HW}$, and regard the last dimension of $A$ as the time dimension and feed it into 3D convolution.

\item \textbf{Step~2} Perform the same operations on $A^\intercal$; reshape time dimensions of two results from $HW$ into $H\times W$ and add them to produce $A^1$.

\item \textbf{Step~3} Crossly feed $A^1$ into two $3\times3\times3$ convolutions, and integrate the results as cross-attention map $A'\in R^{H \times W \times H \times W}$.

\item \textbf{Step~4} Transform $A'$ into cross-attention map $A_q \in R^{H \times W}$ by
\begin{equation}
A_{q}(x_q)=\frac{\exp \left(\frac{\sum_{x_s} A'(x_s, x_q)}{|x_s|} / T\right)}
{\sum_{x'_q}\exp \left(\frac{\sum_{x_s} A'(x_s, x_q)}{|x_s|} / T\right)},
\label{eq.14}
\end{equation}
where $|x_s|=HW$, $T\in (0, 1]$ is the temperature coefficient,
$A_q(x_q)$ represents the average matching probability between two elements in support prototype and query sample at position $x_q$. Similarly, cross attention map $A_s\in R^{H \times W}$ of support prototype can be calculated
by Eq.~(\ref{eq.14}).
\end{itemize}
Finally, $A_q$ and $A_s$ are used to modulate ${F_q}^*$ and ${F_s}^*$, and get the final representations
\begin{equation}
\label{eq.15}
	\begin{cases}
	\hat{F}_q=A_q {F^*_q}, \\
	\hat{F}_s=A_s {F^*_s},
	\end{cases}
\end{equation}
which are used to fuse reliable semantic-related information between them and highlight the relevant regions.
\end{itemize}

\subsection{Learning flow}
\label{sec3.4}

Finally, FicNet uses metric classification for the final representation. It is an end-to-end trainable from scratch. Its loss function and training steps are introduced as follows.
\begin{itemize}[\newItemizeWidth]
\item \noindent\textbf{Metric classification}

For a ``$N$-way $K$-shot" task, $N$ support categories yield $N$ different attention maps. So, one can obtain $N$ different support representations $\{\hat{F}_s^{(i)}\}_{i=1}^N$ and the corresponding query representations $\{\hat{F}_q^{(i)}\}_{i=1}^N$. Then, compute the similarity between support prototypes $\{\hat{F}_s^{(i)}\}_{i=1}^N$
and query representations $\{\hat{F}_q^{(i)}\}_{i=1}^N$ with cosine distance, and then classify query representation into the nearest support category.

\item \textbf{Loss function}

FicNet adopts a single-stage training method, and combines two classification losses for end-to-end learning.
As for module MFN, we classify the query feature $\tilde{F}_q=F_q+F'_q$ by adding an additional full connection layer (the layer FC is only used in the training stage), and calculate the cross entropy loss
\begin{equation*}
L_a=-\log \frac{\exp \left(w_c^{\top} \tilde{F}_q+ b_c\right)}{\sum_{c=1}^{N_{\rm{tr}}} \exp \left(w^{\top}_c \tilde{F}_q + b_c\right)},
\label{eq.16}
\end{equation*}
and use it as the auxiliary loss to guide the learning of CNN-based feature extractor and module MFN, where $N_{\rm{tr}}$ is the total number of categories in train set $D_{\rm{tr}}$, $w^{\top}_c$ and $b_c$ are the weight and bias in the layer FC, respectively.
On the other hand, as for the final representations $\hat{F}_s$, $\hat{F}_q$ output by DCM, we build a metric classification loss $L_c$. Considering that FSFG problem may contain difficult negative samples, we build contrast loss
\begin{equation*}
L_c=-\log \frac{\exp \left(\operatorname{dis}\left(\hat{F}_s^{(i)}, \hat{F}_q^{(i)} \right) / t\right)}{\sum_{i'=1}^N \exp\left(\operatorname{dis}\left(\hat{F}_s^{(i')}, \hat{F}_q^{(i')}\right) / t\right)}
\end{equation*}
and used it as the classification loss, where $\operatorname{dis}(\cdot, \cdot)$ is cosine similarity, the temperature coefficient $t$ is set as 0.2. The classification loss function guides the model to learn the feature space, where the query samples are measured and classified correctly. Finally, the total loss function is
\begin{equation}
L=L_{c}+\mu L_a,
\label{eq.18}
\end{equation}
where $\mu$ is hyper-parameter that balances the loss term.

\item \textbf{Training procedure}

The pseudocodes for implementing the meta-training process
are given in Algorithm~\ref{al.1}.
\end{itemize}

\begin{algorithm}[!htb]
	\caption{Meta-training phase.}
	\label{al.1}
	\SetKwInOut{Input}{Input}\SetKwInOut{Output}{Output}
	\Input{$D_{\rm{tr}}=\{(I_i, y_i), y_i \in C^s\}_{i=1}^{N_{\rm{tr}}}$; learning rate $\alpha$; hyper-parameter $\mu$}
	\Output{Trained model $\varphi \subseteq\{\varphi_{\rm cnn}, \varphi_{\rm mfn}, \varphi_{\rm dcm}\}$}
	\BlankLine
	Randomly initialize model parameters $\varphi_{\rm cnn}$, $\varphi_{\rm mfn}$, $\varphi_{\rm dcm}$; Initialize loss $L=0$\;
	\While{not done}{
	 Sample batch of episodes $T\sim D_{\rm{tr}}$, a task $T_i$ includes a support set $S=\{(I_s, y_s)\}_{s=1}^{NK}$ and a query set $Q=\{(I_q, y_{q})\}_{q=1}^{NP}$ \;
	\For{all $T$}{
	\For{all samples in $T_i$}{
	$F_s, F_q \leftarrow \varphi_{\rm cnn}(I_s, I_q)$ \;
	$F'_s, F'_q, \tilde{F}_s, \tilde{F}_q \leftarrow \varphi_{\rm mfn}(F_s, F_q)$\;
	$ L_a(\varphi) \leftarrow \frac{1}{NP} \sum_{(\tilde{F}_q, y) \in Q} L_a(\tilde{F}_q, y)$ \;
	$\hat{F_s}, \hat{F}_q \leftarrow \varphi_{\rm dcm}({F_s, F'_s},{F_q, F'_q}) $ \;
	$ L_{c}(\varphi) \leftarrow \frac{1}{ \mathit{NP}} \sum_{(\hat{F}_q, y) \in Q} L_{c}(\hat{F_q}, y)$ \;
	$ L_{T_i}(\varphi) \leftarrow L_{c}(\varphi)+L_a(\varphi)$\;
	}
	}
	$ \varphi \leftarrow \varphi-\alpha \frac{1}{|T|} \sum_{T_i} L_{(T_i)}(\varphi)$\;
	}
\end{algorithm}

\section{Performance evaluation}
\label{sec:Experimental}

Four benchmark datasets and the experimental details are first introduced to support
the following performance evaluation of FicNet over some state-of-the-art methods\footnote{The source code of this paper and the test images used are publicly available at \url{https://github.com/ChengqingLi/FicNet}.}.

\subsection{Datasets}
\label{sec.4.1}

FicNet is evaluated on three fine-grained benchmark datasets: ``CUB-200" (200 kinds of birds, 11,788 images) \cite{ye2020few}, ``Stanford Dogs" (120 kinds of dogs, 20,580 images) \cite{khosla2011novel}, ``Stanford Cars" (196 kinds of cars, 16,185 images) \cite{krause20133d}. In addition, we also test the generalization ability of FicNet on a general image classification dataset, mini-ImageNet \cite{vinyals2016matching}. It is a subset of ImageNet \cite{russakovsky2015imagenet} and consists of 60,000 images distributed over 100 different categories. For fair performance comparison, we adopt the data split method used in \cite{li2020bsnet, ruan2021few}. The original images are divided into two disjoint subsets: the auxiliary training set and testing set. For each category in
the auxiliary training set, we divide the data into training dataset and validation dataset. The former is used for training parameters and the latter is employed to monitor the learning process. The details of the mentioned four datasets are shown in Table~\ref{tab.1}, where $N_{\rm total}$ is the original total number of categories in the datasets; $N_{\rm val}$ is the number of categories in validation set.

\begin{table}[!htb]
\caption{Information on the four used benchmark datasets.}
\centering
\begin{tabular}{cccccc}
		\toprule
		\textbf{Dataset} & $N_{\rm total}$ & $N_{\rm{tr}}$ & $N_{\rm val}$ & $N_{\rm test}$ \\
		\midrule
		CUB-200 & 200 & 120 & 30 & 50 \\
		Stanford Dogs & 120 & 70 & 20 & 30 \\
		Stanford Cars & 196 & 130 & 17 & 49 \\
		mini-ImageNet & 100 & 64 & 16 & 20 \\
		\bottomrule
\end{tabular}
\label{tab.1}
\end{table}

\subsection{Experimental configuration}
\label{sec.4.2}

For fair and extensive comparison, we employ ConvNet-4 and ResNet-12 as different backbones followed with the recent FSFG classification works \cite{li2020bsnet, huang2019compare, zhu2020multi, ruan2021few}. All the samples used for training and testing are resized to $84\times 84$.
We set $U=V=5$ for module MFN. As for module DCM, considering memory and time, the size of 3D convolutions is set as $3\times 3\times3$, $T$ of DCA is set as 1 for ``Stanford Dogs" and 2 for others; parameter $\mu$ in Eq.~(\ref{eq.18}) is set as 0.7; SGD is used to optimize the parameters and the learning rate $\alpha$ is initially set as 0.1. We use the episodic training paradigm to train FicNet. For the ``$N$-way $K$-shot" task, we randomly select $N$ categories, and get $K$ arbitrary samples from each category as the support set, and then randomly choose 15 samples as the query set for evaluation. Finally, we report top-1 average classification accuracy with 95\% confidence intervals of 1,200 test episodes sampled randomly. The model is implemented by PyTorch on Tesla K80.

\setlength\tabcolsep{5pt} 
\begin{table*}[!htb]
\caption{Performance comparison of FicNet with typical FSFG image classification methods over three datasets (\%).}
		\centering
		\begin{threeparttable}[b]
		\begin{tabular}{cccccccc}\toprule
			\multirow{3}{*}{Method} & \multirow{3}{*}{Backbone} & \multicolumn{2}{c}{CUB-200} & \multicolumn{2}{c}{Stanford Dogs} & \multicolumn{2}{c}{Stanford Cars}\\
			\cmidrule{3-8}\cmidrule{3-8}
			 & & 5-way 1-shot & 5-way 5-shot & 5-way 1-shot & 5-way 5-shot & 5-way 1-shot & 5-way 5-shot\\ \midrule
			{Matching Net}\cite{vinyals2016matching} & \multirow{16}{*}{ConvNet-4} & 45.30\textpm1.03 & 59.50\textpm1.01 & 35.80\textpm0.99 & 47.50\textpm1.03 & 34.80\textpm0.98 & 44.70\textpm1.03 \\
			{ProtoNet}\cite{snell2017prototypical,cao2022few} & & 37.36\textpm1.00 & 45.28\textpm1.03 & 37.59\textpm1.00 & 48.19\textpm1.03 & 40.90\textpm1.01 & 52.93\textpm1.03 \\
			{MAML}\cite{finn2017model} & & 54.92\textpm0.95 & 73.18\textpm0.77 & 44.64\textpm0.89 & 60.20\textpm0.80 & 46.71\textpm0.89 & 60.73\textpm0.85 \\
			{RelationNet}\cite{sung2018learning} & & 59.58\textpm0.94 & 77.62\textpm0.67 & 43.05\textpm0.86 & 63.42\textpm0.76 & 45.48\textpm0.88 & 60.26\textpm0.85 \\
			{PCM} \cite{wei2019piecewise} & & 42.10\textpm1.96 & 62.48\textpm1.21 & 28.78\textpm2.33 & 46.92\textpm2.00 & 29.63\textpm2.38 & 52.28\textpm1.46 \\
			{GNN} \cite{garcia2017few} & & 51.83\textpm0.98 & 63.69\textpm0.94 & 46.98\textpm0.98 & 62.27\textpm0.95 & 55.85\textpm0.97 & 71.25\textpm0.89 \\
			{CovaMNet} \cite{li2019distribution} & & 52.42\textpm0.76 & 63.76\textpm0.64 & 49.10\textpm0.76 & 63.04\textpm0.65 & 56.65\textpm0.86 & 71.33\textpm0.62 \\
			{DN4} \cite{li2019revisiting} & & 46.84\textpm0.81 & 74.92\textpm0.64 & 45.41\textpm0.76 & 63.51\textpm0.62 & 59.84\textpm0.80 & 88.65\textpm0.44 \\
			{PABN+}\textsubscript{cpt} \cite{huang2019compare} & & 63.36\textpm0.80 & 74.71\textpm0.60 & 45.65\textpm0.71 & 61.24\textpm0.62 & 54.44\textpm0.71 & 67.36\textpm0.61 \\
			LRPABN+\textsubscript{cpt} \cite{huang2020low} & & 63.63\textpm0.77 & 76.06\textpm0.58 & 45.72\textpm0.75 & 60.94\textpm0.66 & 60.28\textpm0.76 & 73.29\textpm0.58 \\
			{ATL-Net} \cite{dong2021learning} & & 60.91\textpm0.91 & 77.05\textpm0.67 & 54.49\textpm0.92 & 73.20\textpm0.69 & 67.95\textpm0.84 & \uwave{89.16\textpm0.48} \\
			BSNet (R\&C) \cite{li2020bsnet} & & 65.89\textpm1.00 & 80.99\textpm 0.63 & 51.06\textpm0.94 & 68.60\textpm0.73 & 54.12\textpm0.96 & 73.47\textpm0.75 \\
			{FOT} \cite{wang2021fine} & & 67.46\textpm0.68 & 83.19\textpm0.43 & 49.32\textpm 0.74 & 68.18\textpm 0.69 & 54.55\textpm0.73 & 73.69\textpm0.65 \\
			{TOAN} \cite{huang2022toan} & & 65.34 \textpm0.75 & {80.43\textpm0.60} & 49.30\textpm0.77 & 67.17\textpm0.49 & 65.90\textpm 0.72 & 84.24\textpm0.48 \\
			{SACN} \cite{ruan2021few} & & \uwave{71.50\textpm1.62} & 79.77\textpm 0.96 & \uwave{64.30\textpm1.74} & 71.65\textpm 1.00 & 68.23\textpm1.69 & 78.70\textpm1.00 \\
			{DLG}\cite{cao2022few} & & {64.77\textpm0.90} & 83.31\textpm0.55 & 47.77\textpm0.86 & 67.07\textpm 0.72 & 62.56\textpm 0.82 & 88.98\textpm0.47 \\
 {FRN+TDM}\cite{Lee_2022_CVPR} & & 70.79\textpm0.24 & \uwave{88.03\textpm0.14} &57.64\textpm0.22 & \uwave{75.77\textpm0.16} & \uwave{68.36\textpm0.22} &86.14\textpm0.13 \\[2pt]
			\hline
			{NDPNet} \cite{zhang2021ndpnet} & \textbackslash & 64.74\textpm0.90 & 80.52\textpm0.63 & 56.21\textpm0.86 & 74.82\textpm0.84 & 71.48\textpm0.89 & \dashuline{91.92\textpm0.91}\\
			BSNet (R\&C) \cite{li2020bsnet} & ResNet-10 & 69.73\textpm0.97 & 82.85\textpm 0.61 & 58.83\textpm0.85 &\dashuline{76.60\textpm0.69} & 66.97\textpm0.99 & 84.09\textpm0.66 \\
			{MML (KL)} \cite{chen2021multi} & ResNet-12 & -- & -- & \dashuline{59.05\textpm0.23} & 75.59\textpm 0.21 & \dashuline{72.43\textpm0.25} & 91.05\textpm0.23 \\
			Meta DeepBDC \cite{XieLLWL22deep}& ResNet-12 &\textbf{83.60\textpm0.40} &\dashuline{92.86\textpm0.21} & -- & -- & -- & -- \\
			{FOT} \cite{wang2021fine} & ResNet-18 & 72.56 \textpm0.65 & 87.22\textpm0.35 & -- & -- & -- & -- \\
			\midrule
			{FicNet} & \multirow{2}{*}{ConvNet-4} & \textbf{75.27\textpm 0.61} & \textbf{88.48\textpm 0.37} & \textbf{64.74\textpm 0.69} & \textbf{79.23\textpm 0.46} & \textbf{77.31\textpm 0.58} & \textbf{89.47\textpm 0.32}\\
			Trend & & $\uparrow$ 3.77 & $\uparrow$ 0.45 & $\uparrow$0.44 & $\uparrow$ 3.46 &
			$\uparrow$ 8.95 & $\uparrow$ 0.31\\ \hline
			{FicNet} & \multirow{2}{*}{ResNet-12} & \dashuline{80.97\textpm 0.57} & \textbf{93.17\textpm 0.32} & \textbf{72.41\textpm 0.64} & \textbf{85.11\textpm 0.37} & \textbf{86.81\textpm 0.47} & \textbf{95.36\textpm 0.22} \\
			Trend & & - & $\uparrow$ 0.31 & $\uparrow$ 8.11 & $\uparrow$ 8.51 & $\uparrow$ 14.38 & $\uparrow$ 3.44 \\
		\bottomrule
		\end{tabular}
 \begin{tablenotes}
 \item[\textbackslash] NDPNet uses a novel feature re-abstraction network as the extractor.

 \item[--] Neither experiment results on the specified settings nor
 open source codes are available for the corresponding method, so accurate comparison results are unavailable.
 \end{tablenotes}
\end{threeparttable}
\label{tab.7}
\end{table*}

\subsection{Comparison with the state-of-the-art methods}

To evaluate the effectiveness of FicNet on fine-grained datasets, we compare it with some state-of-the-art methods, including several specific FSFG methods and four typical FSL methods. Table~\ref{tab.7} lists the comparison results on three fine-grained datasets including ``CUB-200", ``Stanford Dogs" and ``Stanford Cars", with standard ``5-way 1-shot" and ``5-way 5-shot" tasks. Just referenced the previous counterparts, the best results are marked in bold, the second best results with ConvNet-4 and ResNet are marked
in wavy underline and dashed underline, respectively.
Specially, the numbers beside symbol $\uparrow$ indicate improvement percentage of FicNet with respect to the second best result.

From Table~\ref{tab.7}, one can see that when ConvNet-4 is used as the feature extractor, FicNet can achieve improvement 3.77\%, 0.44\% and 8.95\% over the second best method on the three datasets for the ``5-way 1-shot" tasks, respectively. As for the ``5-way 5-shot" task, FicNet achieves accuracy improvement 0.45\%, 3.46\% and 0.31\% over the second best method on the three datasets, respectively. When Resnet-12 is used as the feature extractor, In general, FicNet also significantly outperforms the other comparable methods.
It can obtain an average accuracy 80.97\%, 72.41\% and 86.81\% on the three datasets for the ``5-way 1-shot" task, respectively.
As for the above task, the corresponding accuracy ratios are 93.17\%, 85.11\% and 95.36\%.
Due to the limitation of presentation space, only some typical results are given in Table~\ref{tab.7}. Moreover, the effectiveness of FicNet is unrelated with selected backbone, which verifies its robustness and compatibility.

\subsection{Generalization ability of ResNet over the general FSL tasks}

Most recent image classification methods \cite{chen2019closer, hou2019cross, zhang2020deepemd} use deeper convolutional neural networks as feature extraction backbones, such as ResNet family. To illustrate the generalization ability of FicNet, we perform experiments on mini-ImageNet.
Here, we select ResNet-12 as backbone to perform the comparative experiment on the ``5-way 1-shot" and ``5-way 5-shot" tasks. Table~\ref{tab.8} shows the comparison results, where the best results are marked in bold, and the numbers beside symbol $\uparrow$ indicate improvement of FicNet over the baseline (ProtoNet*).

\setlength\tabcolsep{2pt} 
\begin{table}[!htb]
\caption{Performance comparison of FicNet with typical FSFG image classification methods over mini-ImageNet datasets (\%).}
\centering
		\begin{tabular}{cccc}\toprule
			\multirow{3}{*}{Method} & \multirow{3}{*}{ {Backbone}} & \multicolumn{2}{c}{mini-ImageNet}\\
			\cmidrule{3-4}\cmidrule{3-4}
			 & & 5-way 1-shot & 5-way 5-shot\\ \midrule
		{MatchNet} \cite{vinyals2016matching} & \multirow{16}{*}{ResNet-12} & 63.08\textpm0.80 & 75.99\textpm0.60 \\
		{cosine classifier} \cite{chen2019closer} & & 55.43\textpm0.81 & 77.18\textpm0.61 \\
		{ProtoNet*} \cite{snell2017prototypical} & & 62.39\textpm0.21 & 80.53\textpm0.14 \\
		{TADAM} \cite{chen2019closer} & & 58.50\textpm0.30 & 76.70\textpm0.30 \\
		{CAN} \cite{hou2019cross} & & 63.85\textpm0.48 & 79.44\textpm0.34 \\
		{RFS-simple} \cite{tian2020rethinking} & & 62.02\textpm0.63 & 79.64\textpm 0.44 \\
		{NegMargin} \cite{liu2020negative} & & 63.85\textpm0.81 & 81.57\textpm 0.56 \\
		{DeepEMD} \cite{zhang2020deepemd} & & 65.91\textpm0.82 & 82.41\textpm0.56 \\
		{MATANet} \cite{chen2020multi} & & 60.13\textpm0.81 & 75.42\textpm0.72 \\
		{FEAT} \cite{ye2020few} & & 66.78\textpm0.20 & 82.05\textpm0.14 \\
		{InfoPatch} \cite{liu2021learning} & & 67.67\textpm0.45 & 82.44\textpm 0.31 \\
		{RENet} \cite{kang2021relational} & & 67.60\textpm 0.44 & 82.58\textpm0.30 \\
		{GCLR} \cite{Zhong2022graph} & & 62.53\textpm0.64 & 80.34\textpm0.47 \\
		{MixtFSL} \cite{afrasiyabi2021mixture} & & 63.98\textpm 0.79 & 82.04\textpm0.49 \\
		{MlSo+PN} \cite{zhang2022mlso} & & 66.08\textpm 1.80 & 82.32\textpm 0.66 \\
		{Meta DeepBDC}\cite{XieLLWL22deep}&&67.34\textpm0.43& 84.46\textpm0.28\\
		\hline
		{DC} \cite{lifchitz2019dense} & {ResNet-18} & 62.53\textpm0.19 & 79.77\textpm 0.19 \\
		{S2M2} \cite{mangla2020charting} & {ResNet-34\textsuperscript{+}} & 63.74\textpm0.18 & 79.45\textpm 0.12 \\
		\hline
		{S2M2} \cite{mangla2020charting} & \multirow{5}{*}{WRN-28-10\textsuperscript{+}} & 64.93\textpm0.18 & 83.18\textpm 0.11 \\
		{S2M2(s+f)} \cite{Chen2021few} & & 66.88\textpm0.18 & 84.26\textpm 0.10 \\
		{PPA} \cite{qiao2018few} & & 59.60\textpm 0.41 & 73.74\textpm 0.19 \\
		{PSST} \cite{chen2021pareto} & & 64.16\textpm 0.44 & 80.64\textpm 0.32 \\
		{wDAE-GNN} \cite{gidaris2019generating} & & 61.07\textpm 0.15 & 76.75\textpm 0.11 \\
		\hline
		FicNet & \multirow{2}{*}{ResNet-12} & \textbf{67.82\textpm0.56} & 82.71\textpm 0.40 \\
		{Trend} & & $\uparrow$ 5.43 & $\uparrow$ 2.18\\
		\bottomrule
		\end{tabular}
\label{tab.8}
\end{table}

As can be seen from Table~\ref{tab.8}, even compared with those methods whose backbone network is ResNet-18, ResNet-34 or WRN-28-10, FicNet still demonstrates better performance, where ``+" denotes larger backbones than ResNet-12. As for ``5-way 1-shot" and ``5-way 5-shot" tasks, the classification accuracy of FicNet can reach 67.82\% and 82.71\%, respectively. Especially, the result on ``5 way 1-shot" task is superior to all the other state-of-the-art methods, verifying that FicNet has high stability and strong generalization.

\begin{figure*}[!htb]
\centering
\includegraphics[width=1.4\columnwidth]{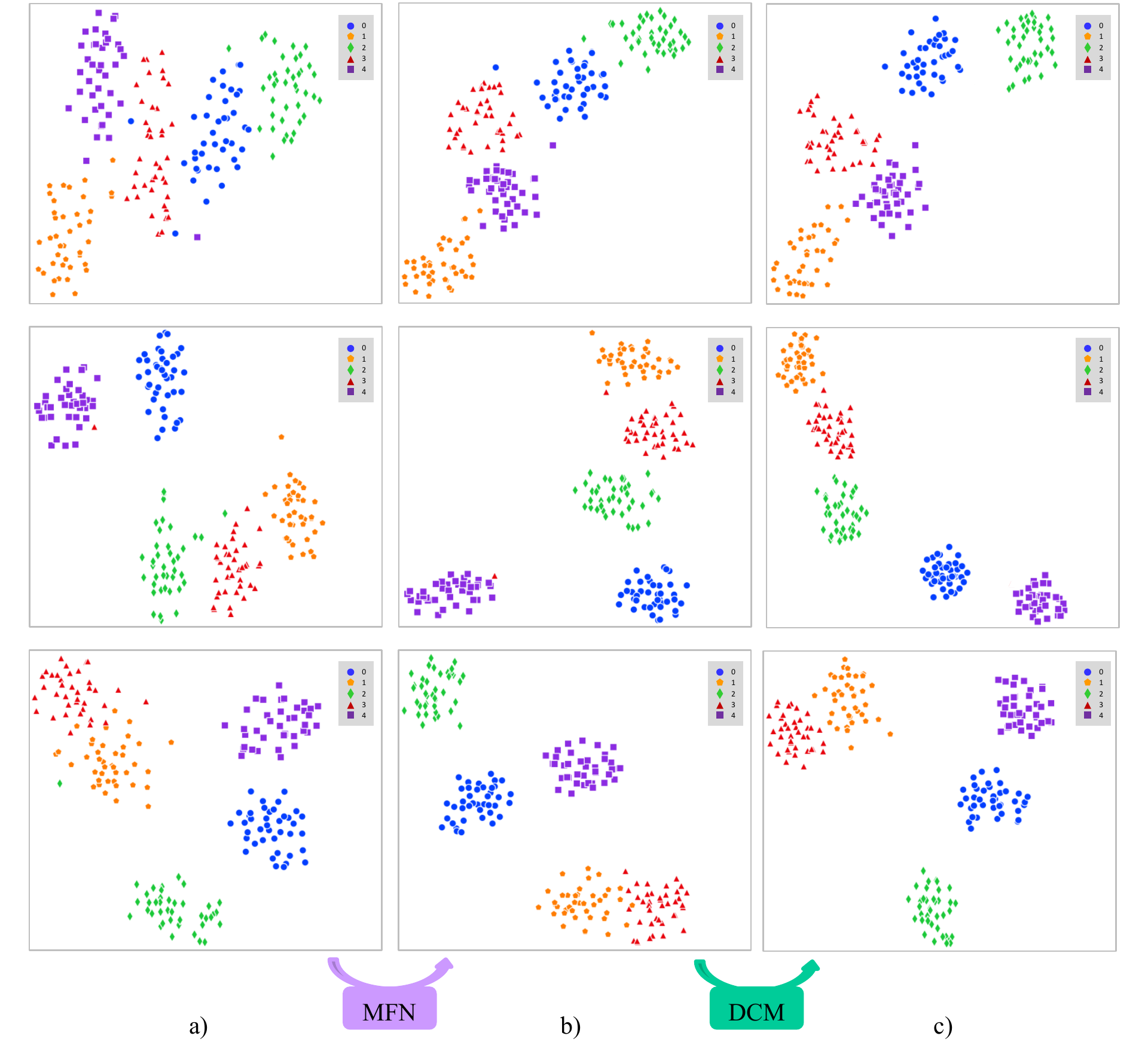}
\caption{t-SNE visualization results of FicNet: a) distribution of basic features; b) distribution of multi-frequency structural features;
c) distribution of final features.}
\label{fig.9}
\end{figure*}

\subsection{Visualization results}

To verify the performance of FicNet in a visual way, we performed t-distributed stochastic neighbor embedding (t-SNE) visualization. We randomly select three ``5-way 5-shot" tasks on dataset ``CUB-200", and each task included 25 support samples and 175 query samples.
The feature distributions of modules MFN and DCM are plotted in Fig.~\ref{fig.9}, where each row represents a ``5-way 5-shot" task, and each column represents the corresponding feature distribution.

From Fig.~\ref{fig.9}a)-b), one can see that the samples belonging to the same category are gradually clustered together and the intra-class distance decreases significantly, which indicates that MFN can learn discriminative class-invariant features. From Fig.~\ref{fig.9}b)-c), one can intuitively observe that the inter-class distance becomes larger and the classification boundary is clearer. In particular, in Fig.~\ref{fig.9}b), some samples distributed at the edge or far from the true category center are prone to be misclassified. After using DCM to do modulation based on global context and task-specific information, the misclassified samples are pulled back to the correct position, which illustrates that DCM can obtain complete and adaptive features that are conductive for good sample matching.

\subsection{Effect of different metric functions}

As we all know, metric function may be a key factor for metric-based learning methods. Therefore, we construct experiments to further explore the performance of FicNet with different metric functions such as Manhattan distance, Euclidean distance and Cosine similarity distance. The results are reported in Table~\ref{tab.metric}, showing that FicNet obtained 80.97\% and 93.17\% on ``1-shot" and ``5-shot" task with cosine similarity metric function, which is the highest accuracy among all comparable metric functions.
Note that the lowest can still achieve 80.29\% and 92.39\% on ``1-shot" and ``5-shot" tasks, which illustrates different metric functions can achieve
similar accuracy. So, the influence of different metric functions is acceptable.
\begin{table}[!htb]
\caption{Classification accuracy (\%) on ``CUB-200" using different metric functions.}
	\centering
	\begin{tabular}{ccc}\toprule
		\multirow{3}{*}{\textbf{Metric functions}} & \multicolumn{2}{c}{5-way Accuracy (\%)}\\
		\cmidrule{2-3}\cmidrule{2-3}
		& 1-shot & 5-shot\\ \midrule
		{Manhattan distance} & 80.29\textpm 0.56 & 92.39\textpm 0.31 \\
		{Euclidean distance} & 80.63\textpm 0.58 & 92.85\textpm 0.32 \\
		{Cosine similarity distance} & \textbf{80.97\textpm 0.57} & \textbf{93.17\textpm 0.32} \\
		\bottomrule
	\end{tabular}
\label{tab.metric}
\end{table}

\section{Ablation Study}
\label{sec:Ablation}

To illustrate the effects of different modules of FicNet, we provide extensive ablation studies using ResNet-12 as backbone.

\subsection{Effectiveness analysis of different modules}

To evaluate the effectiveness of modules MFN and DCM, we conduct experiments with ``5-way 1-shot" and ``5-way 5-shot" tasks on dataset ``CUB-200", where ProtoNet* is used as the baseline.
Table~\ref{tab.2} shows the specific experimental results including accuracy and the number of additional parameters, where symbols ``+" and ``-" indicate the corresponding module is employed and not, respectively.

As can be observed from Table~\ref{tab.2}, compared with the baseline, MFN achieves 5.78\% and 5.84\% accuracy improvements on ``1-shot" and ``5-shot" tasks, respectively.
In contrast, DCM can achieve 5.54\% and 6.23\% improvements on the two tasks, respectively. They also verify the effectiveness of MFN and DCM.
Moreover, combining modules MFN and DCM can achieve classification accuracy 80.97\% and 93.17\% for the two tasks, respectively.
All these results verify that the two modules are orthogonal and complementary, and MFN and DCA are both effective for solving the FSFG problem.

\begin{table}[!htb]
	\caption{Effects of MFN and DCM.}
		\centering
		\begin{tabular}{ccccc}\toprule
			\multirow{1}{*}{MFN} & DCM & \multirow{1}{*}{1-shot} & \multirow{1}{*}{5-shot} & \multirow{1}{*}{\#added params}\\
			\midrule
			- & - & 73.13 & 85.03 & 0K \\
			+ & - & 78.91 & 90.87 & 254.11K \\
			- & + & 78.67 & 91.26 & 551.06K \\
			+ & + & 80.97 ($\uparrow$ 7.84) & 93.17 ($\uparrow$ 8.14) & 805.17K \\
			\bottomrule
		\end{tabular}
\label{tab.2}
\end{table}

\subsection{The performance analysis of MFN}

To further illustrate the effectiveness and generalization of MFN, we provide experiments on the frequency component and its number in MFN, and analyze the performance with other methods.
\begin{itemize}[\newItemizeWidth]
\item \textbf{Influence of the weighted neighborhood}

To illustrate that the weighted neighborhood is reasonable and significant for MFN in spatial domain, we perform corresponding experiments on ``CUB-200". As shown in Table~\ref{weightedneigh}, whether it is ``5-way 1-shot" or ``5-way 5-shot" task, the model performance can be improved to a certain extent after using weighted neighborhood to encode local invariant structure, the improvements are 1.34\% and 0.99\%, respectively. The results verify the necessity and effectiveness of the weighted neighborhood.

\begin{table}[!htb]
\centering
\caption{The accuracy analysis of the weighted neighborhood.}
\begin{tabular}{cccc}
		\hline\noalign{\smallskip}
		\multirow{3}{*}{Weighted neighborhood} & \multicolumn{2}{c}{5-way accuracy (\%)}\\
		\cmidrule{2-3}\cmidrule{2-3}
		& 1-shot & 5-shot\\ \midrule
		w/o &79.63 &92.18\\
		w/ &80.97& 93.17\\
		\noalign{\smallskip}\hline
	\end{tabular}
\label{weightedneigh}
\end{table}

\item \textbf{Influence of individual frequency components}

To study the impact of different frequency components on the performance of MFN, we use only one frequency component of FicNet at a time to evaluate the separate classification accuracy. We divide the 2-D DCT frequency space into $5\times 5$ blocks, and construct 25 experiments on mini-ImageNet with ``5-way 1-shot" task.
The experimental results are shown in Fig.~\ref{fig.6} in a form of $5\times 5$ matrix. As can be seen from Fig.~\ref{fig.6}, the lower frequency components generally have better performance. Their accuracies are different no matter the frequency components are close or far, which show that different frequency components contain different information, and dispel the doubt that the use of multiple frequency components brings information redundancy.
Finally, among all frequency components, the difference scope in classification accuracy is within 2\%. Therefore, even if multiple frequency components are not selected in a targeted manner, it can bring a certain degree of performance improvement, which strongly illustrates that it is an effective innovation to combine information in spatial and frequency domain to encode the features.

\begin{figure}[!htb]
\centering
\includegraphics[width=0.6\BigOneImW]{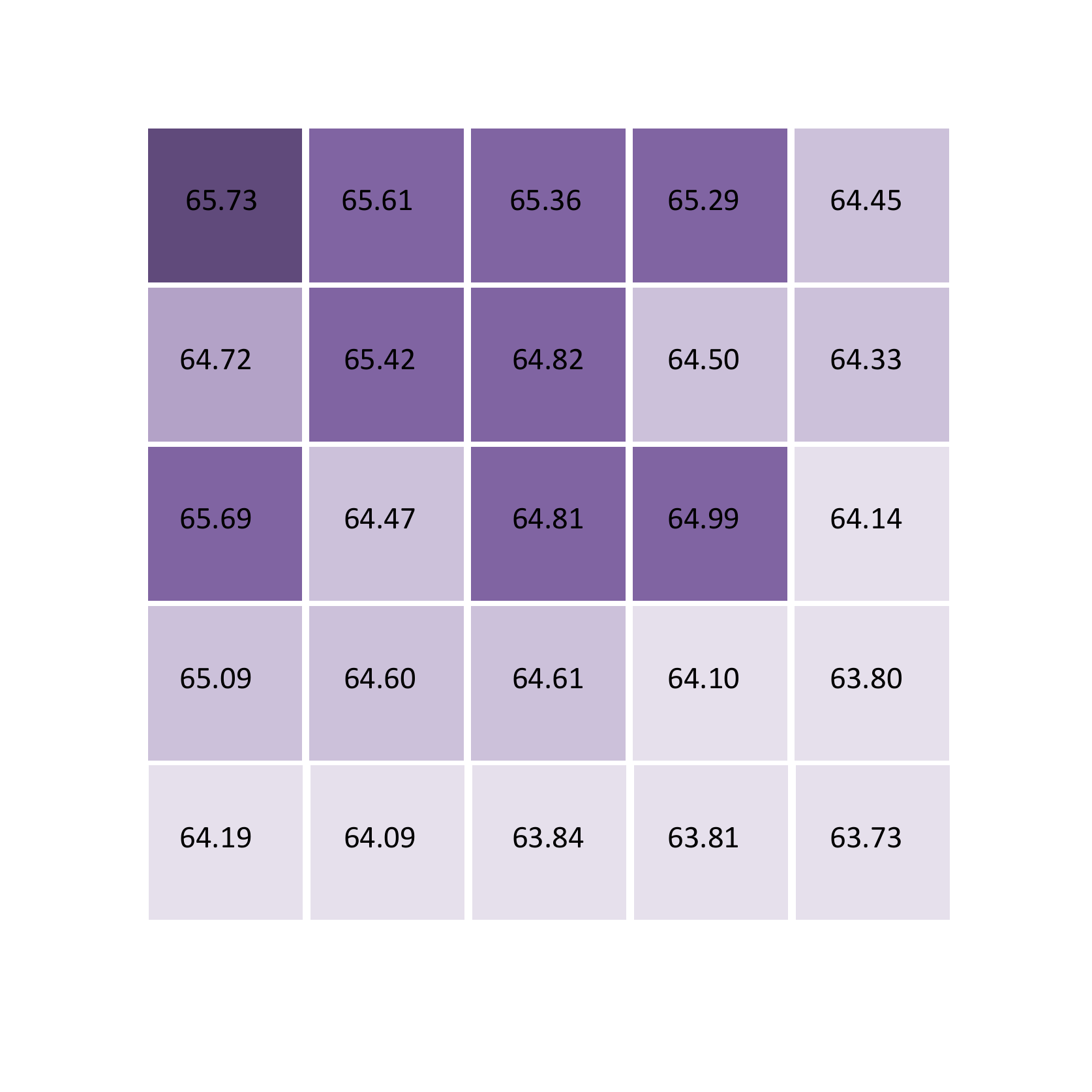}
\caption{Classification accuracy (\%) on mini-ImageNet using different frequency components.}
\label{fig.6}
\end{figure}

\item \textbf{Influence of the number of frequency components}

To further illustrate the influence of the number of frequency components $M$ on model accuracy, we choose $M=1, 2, 4, 8, 12, 16, 25$ to do experiments on dataset ``CUB-200".
Figure~\ref{fig.7} demonstrates that the classification accuracy increases obviously when increasing the number of multiple frequency components from 2 to 12 in the ``5-way 1-shot" and ``5-way 5-shot" tasks.
This is attributed to that different frequency components contain different information, and more information can be extracted from the frequency domain. Moreover, two curves reach peak values 80.97\% and 93.17\% at $M=12$, respectively. This also means that 12 frequency components are sufficient to capture useful features. Therefore, $M$ is set as 12 in FicNet.

\begin{figure}[!htb]
\centering
\input{figures-pdf/fig9.tex}
\caption{Classification accuracy of FicNet using different number of frequency components.}
\label{fig.7}
\end{figure}
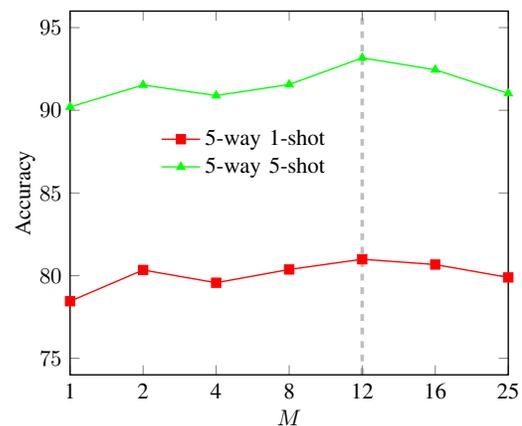

\item \textbf{Effect of MFN in capturing structure representation}

To verify the ability of MFN in capturing the structure representation, we show some test images in ``CUB-200" dataset and the corresponding class activation maps in Fig.~\ref{fig.activemap}. As shown in Fig.~\ref{fig.activemap}c), compared with the activation map of the basic CNN feature extractor in Fig.~\ref{fig.activemap}b), the structure of the target object can be detected after using weighted neighborhood. In Fig.~\ref{fig.activemap}d), when using MFN combined weighted neighborhood with the frequency information, the structure and the contour are more compact and clear, while unrelated information to the structure such as background interference is weaken. It shows that module MFN has
good ability to capture compact and complete structure representation.

\begin{figure}[!htb]
	\centering
	\subfigcapskip=-3pt 
 \subfigure[]{
 \includegraphics[width=0.2\linewidth]{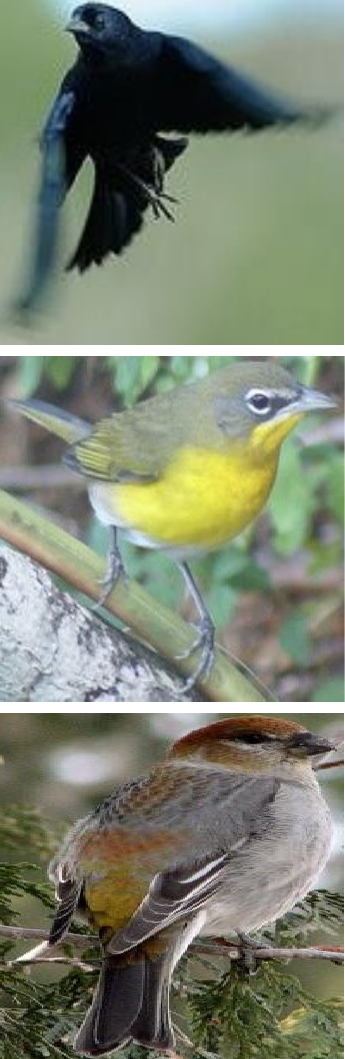}}
 \hspace{-0.3cm}
 \subfigure[]{
 \includegraphics[width=0.2\linewidth]{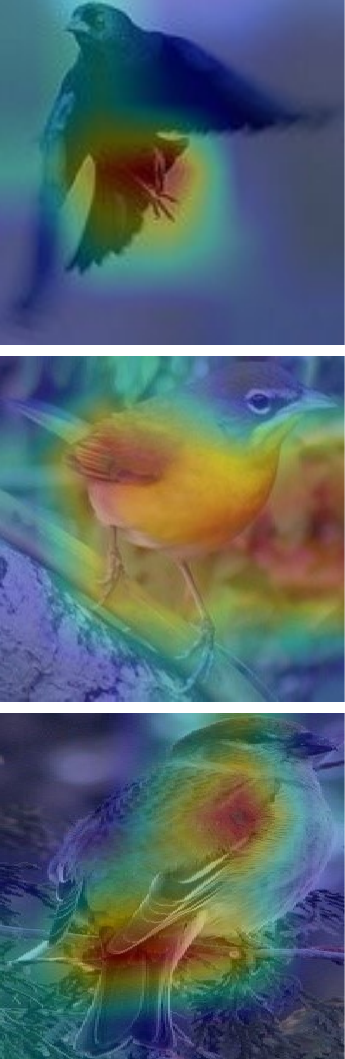}}
 \hspace{-0.3cm}
 \subfigure[]{
 \includegraphics[width=0.2\linewidth]{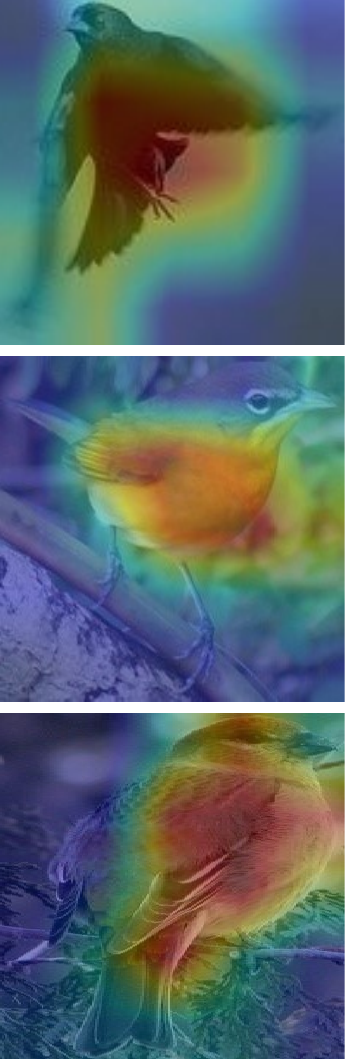}}
 \hspace{-0.3cm}
 \subfigure[]{
 \includegraphics[width=0.2\linewidth]{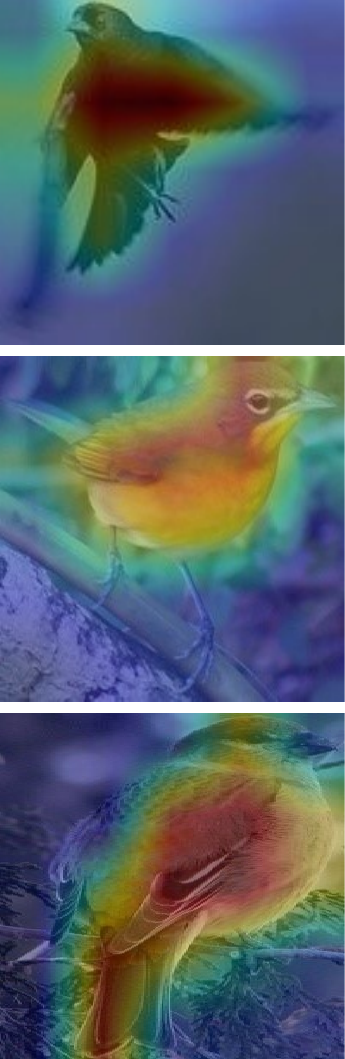}}
 \vspace{-0.15cm}
	\caption{The capturing performance of structure representation through various class activation maps: (a) origin images; (b) basic CNN feature extractor; (c) weighted neighborhood; (d) MFN.}
\label{fig.activemap}
\end{figure}

\item \textbf{Boosting other classification methods by MFN}

We explore generalization and applicability of MFN by combining it with other models. Here, we combine
MFN with
ProtoNet* in \cite{snell2017prototypical} and CAN in \cite{hou2019cross} to illustrate the performance. Table~\ref{tab.3} lists the classification accuracy of datasets ``CUB-200" and mini-ImageNet with ``5-way 1-shot" task.
As can be seen from Table~\ref{tab.3}, MFN improves the accuracy of ProtoNet* by 5.78\% on dataset ``CUB-200" and 4.43\% on dataset mini-ImageNet. It also obtains
 1.96\% and 3.08\% improvement over the results obtained with CAN on
the two datasets, respectively.
All these results convince us that MFN can boost the performance of other methods significantly.

\begin{table}[!htb]
\caption{Results of combining MFN with other methods.}
\centering
\begin{tabular}{lcc}\toprule
		method & {``CUB-200"} & {mini-ImageNet}\\
		\midrule
		ProtoNet* \cite{snell2017prototypical} & 73.13 & 62.39\\
		ProtoNet*+MFN & 78.91 & 66.82\\
		CAN \cite{hou2019cross} & 77.22 & 63.85\\
		CAN+MFN & 79.18 & 66.93\\
		\bottomrule
\end{tabular}
\label{tab.3}
\end{table}

\item \textbf{Comparing MFN with other similarity-based modules}

To show the effectiveness of MFN, we compare it with other similarity-based modules by adding them to the baseline model ProtoNet*. The module SCE in \cite{huang2019dynamic} calculates the cosine similarity between the referenced position and its adjacent ones, and connects their similarity in a fixed order of channel dimension. On the contrary, MFN does not limit observation of different neighboring relation.
SCR in \cite{kang2021relational} uses self-similarity to learn the relation pattern of ``how each feature is related to its neighbors". However, it does not consider that different neighbors have different importance. MFN uses weighted neighbors and integrates frequency information to capture more compact structural patterns. As shown in Table~\ref{tab.4}, compared with SCE, the improvement of the MFN is 3.63\% on mini-ImageNet, and it outperforms all the other similarity-based modules. The results verify the effectiveness of MFN.
\end{itemize}

\begin{table}[!htb]
	\caption{Comparison with other similarity-based modules.}
	\centering
	\begin{tabular}{lcc}\toprule
		method & CUB-200 & mini-ImageNet\\
		\midrule
		Baseline & 73.13 & 62.39\\
		Baseline+SCE \cite{huang2019dynamic} & 77.85 & 63.19\\
		Baseline+SCR \cite{kang2021relational} & 78.09 & 66.18\\
		Baseline+MFN & 78.91 & 66.82\\
		\bottomrule
	\end{tabular}
\label{tab.4}
\end{table}

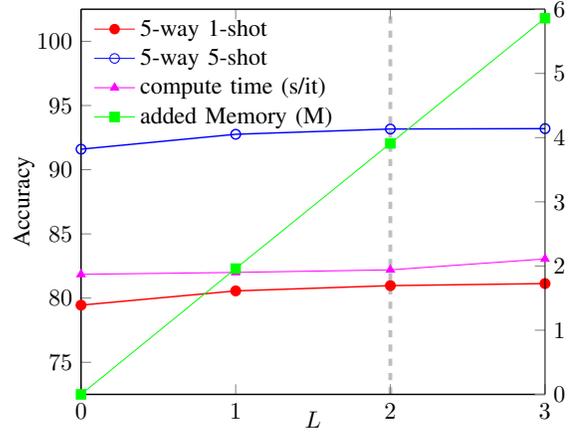
\begin{figure}[!htb]
\centering
\input{figures-pdf/fig11.tex}
\caption{Experimental results of FicNet using different number of crisscross operations.}
\label{fig.8}
\end{figure}

\subsection{The performance analysis of DCM}
\label{sec.5.3}

To deeply illustrate the performance of DCM, we study how to select the number of crisscross operations of BCC and analyze the performance of BCC and DCA for the modulation.

\begin{itemize}[\newItemizeWidth]

\item \textbf{Influence of the number of crisscross operation}

To show the effect of using different number of crisscross operation in BCC, Fig.~\ref{fig.8} depicts the variation curve of classification accuracy, memory and time with the number $L$.
As shown in Fig.~\ref{fig.8}, after adding one crisscross operation into FicNet, the performances of ``1-shot" and ``5-shot" tasks are improved by 1.11\% and 1.16\%, respectively. This can effectively demonstrates importance of supplementing context information. Then, if increase number $L$ from 1 to 2, that is, using BCC component, the model can improve the performance under both tasks by another 0.41\%.
Furthermore, in Fig~\ref{fig.8}, we also observe the memory occupancy and computation time continue to increase with respect to $L$.
Therefore, to balance the performance and computational resources, $L=2$
is optimal for FicNet.

\item \textbf{Effectiveness analysis of BCC and DCA of DCM}

To study the effect of BCC and DCA of DCM, we employ ablation experiments on dataset ``CUB-200" for ``1-shot" and ``5-shot" tasks. Note that if the module is employed to FicNet, it is represented with ``+" and ``-" otherwise. The results are illustrated in Table~\ref{tab.6}, which show that BCC achieves 0.27\% and 0.15\% improvements
over the baseline for ``1-shot" and ``5-shot" tasks.
In contrast, DCA obtains 1.02\% and 0.45\% improvements on the two tasks, respectively. Moreover, combining BCC with DCA can achieve classification accuracy 2.06\% and 2.3\% improvement over the baseline for the two tasks, respectively. All these verify that the combination of BCC and DCA can further improve the classification accuracy and bring benefits to the model, which also shows that it is necessary to integrate the global context information into the embedding process.

\setlength\tabcolsep{7pt} 
\begin{table}[!htb]
\caption{Effect of BCC and DCA of module DCM}
		\centering
		\begin{tabular}{cccc}\toprule
		\multicolumn{2}{c}{DCM} & \multirow{3}{*}{1-shot} & \multirow{3}{*}{5-shot}\\
			\cmidrule{1-2} \cmidrule{1-2}
			BCC & DCA & & \\\midrule
			 - & - & 78.91 & 90.87\\
			 + & - & 79.18 & 91.02\\
			- & + & 79.93 & 91.32\\
			 + & + & 80.97 & 93.17\\
			\bottomrule
		\end{tabular}
\label{tab.6}
\end{table}

\item \textbf{Comparison of DCA with other cross-attention modules}

To further analyze the performance of DCA component, we construct experiments with DCA and other cross-attention modules under the same experimental setting. Similar to DCA component, CAN in \cite{hou2019cross} and the cross-correlational attention (CCA) in RENet\cite{kang2021relational} also focus on task-specific information by using cross attention, and the former directly averages 4D cross-correlation between images into 2D tensor, the latter learns reliable co-attention via 4D convolutional filter. Compared with them, DCA uses 3D double-cross convolutions, which crosswises and progressively refines the 4D cross-correlation tensor while maintaining the details of geometric space. The DCM module in FicNet not only contains DCA component, but also focuses on context information with BCC component. The comparison results are shown in Table~\ref{tab.5}. As shown in the first three rows of Table~\ref{tab.5}, DCA component performs best. Especially when replacing CAN and CCA with DCA component, the classification accuracy is improved by 0.71\% ,0.34\% and 1.18\%, 0.74\% on ``CUB-200" and mini-ImageNet, respectively. The results illustrate that DCA can avoid the collapse of key details caused by CAN, and obtain more precise task-level attention than CCA. In addition, when replacing CCA with DCM module, the classification accuracy is improved by 1.08\% and 1.97\% on ``CUB-200" and mini-ImageNet respectively. And the last two rows of Table~\ref{tab.5} demonstrate that the classification accuracy can be improved after adding BCC components. The results also explain that context and task information can both supplement each other in the modulation branch.

\begin{table}[!htb]
	\caption{Comparison results with other cross-attention modules.}
	\centering
	\begin{tabular}{lcc} \toprule
		method & {``CUB-200"} & mini-ImageNet\\
		\midrule
		CAN & 77.22 & 63.85\\
		CCA & 77.59 & 64.29\\
		DCA & 77.93 & 65.03\\
		DCM & 78.67 & 66.26\\
		MFN+CAN & 79.18 & 66.93\\
		MFN+DCA & 79.93 & 67.21\\
		MFN+BCC+CAN & 80.02 & 67.25\\
		MFN+BCC+DCA & 80.97 & 67.82\\
		\bottomrule
	\end{tabular}
\label{tab.5}
\end{table}
\end{itemize}

\section{Conclusion}

In this paper, we designed a high-accuracy few-shot fine-grained image classification method by capturing and modulating with two modules MFN and DCM. The former captured multi-frequency structural representation from spatial and frequency information, and could bring a certain degree of performance improvement. The latter modulated feature from both global context and task-specific information. The proposed method did not require bounding box or part annotation in new representation for metric classification.
Extensive experiments demonstrated that it outperforms the previous state-of-the-art counterparts on some fine-grained datasets.
However, its performance may be less significant when dealing with richer datasets. Extending its generalization and effectiveness to efficiently deal with more complex scenes and computer vision tasks, such as salient object detection, deserves further exploring.

\bibliographystyle{IEEEtran-doi}
\bibliography{FicNet}

\end{document}

%% file: figures-pdf/fig9.tex
\begin{tikzpicture}[scale=0.85]
\begin{axis}[
enlarge y limits = 0.05,
symbolic x coords={1,2,4,8,12,16,25},
xtick = data,
ylabel = Accuracy,
ylabel style = {yshift=-0.5em}, 
xlabel = $M$,
xlabel style = {yshift=0.2em},
xmin=1, xmax=25, ymin = 75, ymax = 95,
minor xtick={12}, 
grid=minor,
line width = 0.6pt,
tick style = {line width = 0.1pt},
grid style = {ultra thick,dashed},
legend style= {anchor=north,at={(0.4,0.7)},legend columns=1,legend cell align=left, align=left, fill=none, draw=none},
]
\addplot[color=red, mark=square*, mark options={solid, red}]
  table[row sep=crcr]{%
1	78.45\\
2	80.34\\
4	79.57\\
8	80.37\\
12	80.99\\
16	80.67\\
25	79.89\\
};
\addlegendentry{5-way 1-shot}
\addplot[color=green, mark=triangle*, mark options={solid, green}]
  table[row sep=crcr]{%
1	90.2 \\
2	91.53\\
4	90.89\\
8	91.56\\
12	93.17\\
16	92.45\\
25	91.02\\
};
\addlegendentry{5-way 5-shot}
\end{axis}
\end{tikzpicture}

%% file: figures-pdf/fig11.tex
\begin{tikzpicture}[scale=0.9]
\pgfplotsset{%
xmin=0, xmax=3,
xtick = {0, 1, 2, 3},
}
\begin{axis}[
xlabel={$L$},
xlabel style = {yshift=1em},
ymin=75,
ymax=100,
ytick={75, 80, 85, 90, 95, 100},
axis y line*=left,
enlarge y limits = 0.1,
ylabel = {Accuracy},
ylabel style = {yshift=-0.6em},
minor xtick={2}, 
grid=minor,
line width = 0.6pt,
tick style = {line width = 0.1pt},
grid style = {ultra thick, dashed},
]
\addplot[color=red, mark=*, mark options={solid, red}]
table[row sep=crcr]{%
0	79.45\\
1	80.56\\
2	80.97\\
3	81.13\\
};
\label{plot_one}
\addplot[color=blue, mark=o, mark options={solid, blue}]
table[row sep=crcr]{%
0	91.6  \\
1	92.76 \\
2	93.17 \\
3	93.2  \\
};
\label{plot_two}
\end{axis}

\begin{axis}
[
ymin = 0,ymax = 6,
ytick = {0,1,2,3,4,5,6},
axis x line = none,
axis y line* = right,
ylabel near ticks,
legend style={at={(0.29, 1.00)}, legend columns=1, anchor=north, legend cell align=left, align=left, fill=none, draw=none}
]

\addlegendimage{/pgfplots/refstyle=plot_one}
\addlegendentry{5-way 1-shot}
\addlegendimage{/pgfplots/refstyle=plot_two}
\addlegendentry{5-way 5-shot}
\addplot[color=mycolor1, mark=triangle*, mark options={solid, mycolor1}]
table[row sep=crcr]{%
0	1.87 \\
1	1.9  \\
2	1.94 \\
3	2.11 \\
};

\addlegendentry{compute time (s/it)}
\addplot[color=green, mark=square*, mark options={solid, green}]
table[row sep=crcr]{%
0	0    \\
1	1.96 \\
2	3.91 \\
3	5.86 \\
};
\addlegendentry{added Memory (M)}
\end{axis}
\end{tikzpicture}